\documentclass{sig-alternate}
\usepackage{amssymb}
\usepackage{xspace}
\usepackage{latexsym}
\usepackage{url}
\usepackage{subfigure}

\begin{document}

\conferenceinfo{CARS-2009,}{October 25, 2009, New York, NY, USA.\\ Copyright is held by the author/owner(s).}


\title{Using Contextual Information as Virtual Items on Top-N Recommender Systems}

%
%
%
%
%

\numberofauthors{3} 
%
\author{
%
%
\alignauthor
Marcos A. Domingues\\
       \affaddr{Fac. of Science, U. Porto}\\
       \affaddr{LIAAD-INESC Porto L.A.}\\
       \affaddr{Portugal}\\
       \email{marcos@liaad.up.pt}
 \alignauthor
 Al\'{i}pio M\'{a}rio Jorge\\
       \affaddr{Fac. of Science, U. Porto}\\
        \affaddr{LIAAD-INESC Porto L.A.}\\
        \affaddr{Portugal}\\
        \email{amjorge@fc.up.pt}
 \alignauthor Carlos Soares\\
       \affaddr{Fac. of Economics, U. Porto}\\
        \affaddr{LIAAD-INESC Porto L.A.}\\
        \affaddr{Portugal}\\
        \email{csoares@fep.up.pt}
}

\maketitle
\begin{abstract}
Traditionally, recommender systems for the Web deal with applications that have
two dimensions, users and items. Bas\-ed on access logs that
relate these dimensions, a recommendation model can be built and used to
identify a set of $N$ items that will be of interest to a certain user.
In this paper we propose a method to complement the information in the access
logs with contextual information without changing the recommendation algorithm. The method consists in representing
context as virtual items. We empirically test this
method with two top-$N$ recommender systems, an item-based collaborative
filtering technique and association rules, on three data sets. The results show
that our method is able to take advantage of
the context (new dimensions) when it is informative.
\end{abstract}




\section{Introduction}
\label{sec:int}
Most Web sites offer a large number of information resources to their users.
Finding relevant content has, thus, become a challenge for users. Recommender
systems have emerged in response to this problem. A recommender system for a
Web site receives (implicit or explicit) information about users and their behavior
and recommends items that are likely to fit his/her needs~\cite{sarwar:00}.

Recommender models for Web personalization can be built from
the historical record of accesses to a site, where one access is a pair $<user\_id,item>$.
Each access is interpreted as a rating of $1$ given by the user to the item.
However, other dimensions, such as time and location, can add contextual
information and improve the accuracy of recommendations. For instance, the type of books that a user looks for in Amazon during work hours is probably different from the books searched for during leisure hours.

According to~\cite{palmisano:08}, the idea that contextual information is
important when predicting customer behavior is not new. Many Web sites are supported by Content Management Systems (CMS), that
often store much contextual information. However, this is not true in all cases
and, additionally, getting information that is really relevant for
recommendation is a hard task in many applications~\cite{gorgoglione:06}. 
Adomavicius et al.~\cite{adomavicius:05a} have investigated the use of context
for rating estimation in multidimensional recommender systems. Palmisano
et al.~\cite{palmisano:08} have used contextual information to improve the
predictive modeling of customer's behavior. Both authors have developed a
special-purpose browser to obtain rich contextual information.

In this paper we exploit how contextual information can be used to improve the accuracy of Top-$N$ Recommender Systems. Existing contextual recommender
systems typically use contextual information as a label for segmenting/filtering
sessions, using them to build the recommendation model
(e.g.,~\cite{adomavicius:05a,palmisano:08}). We follow an alternative approach,
which uses the contextual attribute as a virtual item. This means that it is
treated as an ordinary item for building the recommendation model, which has
the advantage of allowing the use of existing recommendation algorithms.
As our contextual information are obtained from multidimensional data,
we have called our approach \textbf{DaVI} (\emph{Dimensions as Virtual Items}).
Instead of a special-purpose browser~\cite{adomavicius:05a,palmisano:08},
we collect the multidimensional data from Web access logs and from attributes
stored in databases of the Web sites. We have empirically tested our approach
with two recommendation techniques, item-based collaborative filtering and
association rules, to assess the effect of adding context on the accuracy of
traditional Web recommender systems. We present results obtained on three data sets.

In the following section, we present the contextual information used in our experiments.
Next, we describe the recommendation techniques and the approach proposed.
Then, we discuss results and present conclusions and future work.

\section{Contextual Information}
\label{sec:multi}
There are many definitions of context in the literature depending on the field of
application and the available customer data~\cite{palmisano:08}.
In this paper, context is defined as any information that can be used to characterize
the situation of an entity~\cite{dey:01}. Here an entity is an access to an item/Web page by a user.

A critical issue is how to obtain the rich contextual
information~\cite{dey:01a}.
In some circumstances, context is explicit, such as a person informing a movie
recommender system where he/she wants to watch a movie. On the other hand, the
contextual information can also be inferred from Web access data. For example,
we can observe if a person bought an item, from an e-commerce Web site, on a weekday
or a weekend, from the Web access logs.

Besides general contextual information that can be obtained from access logs, we
may use domain-specific information, that is typically collected from the CMS.
For example, if an item represents an access to a music, the genre of the music
can be used as a dimension of contextual information.

In Table~\ref{table:dimensions} we present the dimensions/contextual information
considered in the experiments presented in this paper. The first group of
contextual information was obtained by pre-processing Web access logs. The second
group was collected from the CMS of a Web site of Portuguese Music\footnote
{http://www.palcoprincipal.pt.} used in this study. The last group refers to a
public data
 set\footnote{http://archive.ics.uci.edu/ml/datasets/Entree+Chicago+\\Recommendation+Data.}
 that contains a record of user interactions with the Entree Chicago restaurant
 recommender system. All the information
is stored in a data warehouse that was specifically designed for modeling Web sites~\cite{domingues:07}.

\begin{table}[h]
\caption{Contextual information}
\label{table:dimensions} \footnotesize
\centering                          
\begin{tabular}{p {1.7 cm } p {6 cm }}
\hline
\textbf{Context} & \textbf{Description}\\
\hline \hline
\textit{day} & Day of each access (from 01 to 31).\\
\textit{month} & Month of each access (from 01 to 12).\\
\textit{week\_day} & Week day of each access (from Monday to Sunday).\\
\textit{work\_day} & If the accesses were made during the week (from Monday to Friday) or weekend (Saturday or Sunday).\\
\textit{hour} & Hour of each access (from 01 to 24).\\
\textit{work\_hour} & If the accesses were made during the working time (from 8 a.m. to 6 p.m.) or not.\\
\textit{location} & Location where the accesses were made (country).\\
\hline
\textit{music\_genre} & The genre of a music. There are 45 different musical genres, for instance, pop, rock, jazz, and so forth.\\
\textit{band} & The band which plays a music. There are 2296 different bands in our music recommendation data sets.\\
\textit{instrumental} & If a music is instrumental or not.\\
\hline
\textit{intention} & The intention of navigation in a restaurant recommendation system (for example, the search for a restaurant cheaper, closer, more traditional, more creative, and so forth). There are 9 different intentions of navigation in our experiments.\\
\hline
\end{tabular}
\end{table}

\section{Recommender Systems}
\label{sec:recsys} A recommender system for the Web typically
outputs an ordered list of recommendations, given a trail of recent Web page requests. Historical information about the behavior of the users of the site and the current
session are used to suggest certain pages or services,
or even the purchase of certain products~\cite{sarwar:00}. In the context of the Web, a session can be abstracted to
a set of pairs $<user\_id, item>$, recorded at moments close in time, with
the same $user\_id$.

Usually a recommender system is divided into a two-stage
process~\cite{anand:03}. The first stage is carried out offline.
Data representing the behavior of users of the Web site, which was
previously collected are mined and a model is generated for use in future online interactions. The second stage is carried out in
real-time with a new user interacting with the Web site. Data from
the current user session are used as input by the model to generate
a list of $N$ recommendations. A number of algorithms have been used for
offline model building~\cite{anand:03}, including Collaborative Filtering, Item-Based Collaborative Filtering, Association Rules and Markov Models. In this section we briefly describe the two algorithms used in this work, Item-Based Collaborative Filtering and Association Rules, and how we have applied \textbf{DaVI} (\emph{Dimensions as Virtual Items}) on these algorithms.

\subsection{Item-Based Collaborative Filt\-ering}
\label{subsec:cf}
Item-based collaborative filtering (CF) analyzes stored accesses (grouped in sessions) to identify relations between the items in the set $I$, which contains all items of a Web site~\cite{karypis:01}. The recommendation model is a matrix representing the similarities between all pairs of items, according to a chosen similarity measure. An abstract representation of a similarity matrix is shown as Table~\ref{table:sim_matrix}.

\begin{table}[!h]
\caption{Item-item similarity matrix.}
\label{table:sim_matrix} \footnotesize
\centering                          
\begin{tabular}{|c|c|c|c|c|}
\hline
 & \textbf{$i_1$} & \textbf{$i_2$} & $\cdots$ & \textbf{$i_k$}\\
\hline
\textbf{$i_1$} & $1$ & $sim(i_1,i_2)$ & $\cdots$ & $sim(i_1,i_k)$\\
\hline
\textbf{$i_2$} & $sim(i_2,i_1)$ & $1$ & $\cdots$ & $sim(i_2,i_k)$\\
\hline
$\cdots$ & $\cdots$ & $\cdots$ & $1$ & $\cdots$\\
\hline
\textbf{$i_k$} & $sim(i_k,i_1)$ & $sim(i_k,i_2)$ & $\cdots$ & $1$\\
\hline
\end{tabular}
\end{table}

In Table~\ref{table:sim_matrix}, each item $i \in I$ is an accessed page. The similarity measure used here is the cosine angle, defined by

\begin{center}
$sim(i_{k_1},i_{k_2}) = cos(\overrightarrow{i_{k_1}},\overrightarrow{i_{k_2}}) =
\frac{\overrightarrow{i_{k_1}}.\overrightarrow{i_{k_2}}}{||\overrightarrow{i_{k_1}}||*||\overrightarrow{i_{k_2}}||},$
\end{center}

\noindent{where $\overrightarrow{i_{k_1}}$ and $\overrightarrow{i_{k_2}}$ are binary vectors with as many positions as existing users. The value $1$ means that the users accessed the respective item/page. The value $0$ is the opposite. The ``.'' denotes the dot-product of the two vectors.}

Given a user who accessed the set of items $O \subseteq I$, the model generates a
recommendation by selecting the $N$ which are the most similar to the items in the
set $O$. Here, the similarity for each item $i \notin O$ is given by the weighted average of its nearest neighbors with respect to their presence in the set $O$.



\begin{table}[!ht]
\caption{Similarity matrix with the contextual information $day$.}
\label{table:sim_matrix_day} \footnotesize
\centering                          
\begin{tabular}{|c|c|c|c|c|c|}
\hline
               & \textbf{$i_1$} & $\cdots$ & \textbf{$i_k$} & $\cdots$\\
\hline
\textbf{$i_1$} & $1$ & $\cdots$ & $sim(i_1,i_k)$ & $\cdots$\\
\hline
\textbf{$\cdots$} & $\cdots$ & $1$ & $\cdots$ & $\cdots$\\
\hline
\textbf{$i_k$} & $sim(i_k,i_1)$ & $\cdots$ & $1$ & $\cdots$\\
\hline
\textbf{$d_1$} & $sim(d_1,i_1)$ & $\cdots$ & $sim(d_1,i_k)$ & $\cdots$\\
\hline
$\cdots$ & $\cdots$ & $\cdots$ & $\cdots$ & $\cdots$ \\
\hline
\textbf{$d_v$} & $sim(d_v,i_1)$ & $\cdots$ & $sim(d_v,i_k)$ & $\cdots$\\
\hline
\end{tabular}
\end{table}

When we apply \textbf{DaVI} on the item-based
collaborative filtering algorithm, it treats the contextual attributes as new items (virtual items) in the data set. This means that it adds a new row and column for each different value of the context to the former similarity matrix and calculates the corresponding similarity values, among the values of the context and the other items, as presented previously. A representation of a similarity matrix with contextual information $day = \{d_1,d_2,\cdots,d_v\}$ is shown on Table~\ref{table:sim_matrix_day}. Here, an item can be a page or a possible value for the context day ($1$ to $31$).  Although the contextual information is used in the models, only pages are recommended. The recommendations will be the set of pages that are most similar to a given set of observable items $O \subseteq \{I \cup day\}$. The rationale behind this approach is that the similarity between a given item and a given day (for example)
is higher if the item tends to be accessed on that day of the month. This way, the relation between items and the context is captured. When a recommendation is made for an active session, the value of the context on that particular session (e.g., the day of the month the active session is taking place) is used to provide the contextual information.

\subsection{Based on Association Rules}
\label{subsec:ar} 
A recommendation model $M$ based on association rules (AR) is a set of rules $R$,
each of them with the form $A \rightarrow B$, where $A$ and $B$ are sets of
items. Each AR is characterized by their support and
confidence~\cite{agrawal:94}. The model is generated from a set of Web sessions,
consisting of a set of pairs $<id,item>$ with the same $id$, where $id$ and
$item$ identify the user and the accessed page. Given a set of observable items
$O$, the set of rules $R$ is used to recommend a set of items/pages $Recs$, as
follows:

%
%

\begin{center}
$Recs = \{consequent(r_i)|r_i \in M$ and $antecedent(r_i) \subseteq O$ and $consequent(r_i) \notin O \}$.
\end{center}

\noindent{To obtain the top $N$ recommendations, we select from $Recs$ the
distinct recommendations corresponding to the rules with the highest confidence.
In our work we use the
\emph{\textbf{Caren}}\footnote{http://www.di.uminho.pt/\~{}pja/class/caren.html.}
association rules generator.}

Extending AR to handle contextual information by applying
\textbf{DaVI}, simply consists of including extra
pairs user-item into the former set of sessions. For example, to use the
dimension $day$, we add a pair $<id,day=value>$ to the respective session with
tag $id$, where $day=value$ represents the day of the month when the session $id$
occurred. The set of augmented sessions are used as input to the recommendation
algorithms.
The rules built will include both actual items and virtual items on the
antecedent and only actual items on the consequent. Given an active session
occurring on day $x$, the set of observables $O$ includes the items in the active
session and the virtual items (e.g. $day=x$).

Notice that \textbf{DaVI} does not modify the recommendation algorithms. It just inserts the contextual information as virtual items in the data sets. Thus we can easily extend \textbf{DaVI} to other recommendation methods.

\section{Empirical Evaluation}
\label{sec:expe} In this section we evaluate how \textbf{DaVI} can improve the
accuracy of the recommendation algorithms presented in Sections~\ref{subsec:cf}
and~\ref{subsec:ar}.


\subsection{Experimental Setup}
\label{subsec:set}
The evaluation is carried out on three different data sets
(Table~\ref{table:datasets}). The \emph{Listener} data set contains accesses to
songs in the music Web site mentioned earlier. The \emph{Playlist} data set
represents the set of songs explicitly selected by users of the same site for
their individual playlists. \emph{Entree} is a public data set that contains a
record of user interactions with the Entree Chicago restaurant recommender
system.

\begin{table}[ht]
\caption{Characteristics of the data sets}
\label{table:datasets} \footnotesize
\centering                          
\begin{tabular}{cccc}
\hline
\textbf{Data sets} & \textbf{\# Accesses} & \textbf{$\neq$ Items} & \textbf{$\neq$ Users}\\
\hline \hline
\textit{Listener} & 62208 & 6428 & 9740\\
\textit{Playlist} & 37022 & 5428 & 4417\\
\textit{Entree} & 149849 & 639 & 31440\\
\hline
\end{tabular}
\end{table}

To measure the accuracy of the recommender systems we use the All
But One protocol~\cite{breese:98}. In this protocol,
the sessions in the data set are split randomly into train and test. In our case, 80\% for training and 20\% for testing. The training set is used to generate the recommendation model (similarity matrix or association rules). For each session in the
test set we randomly delete one pair $<id,item>$, referred to as \textit{hidden} item.
The remaining pairs represent the set of observables, $O$, based on which the recommendation is made.

The model is evaluated by comparing, for each session in the test set,
the set of recommendations it makes (\textit{Rec}), given the set of
observables, $O$, against the hidden item. The set of recommendations
${rec_1,rec_2,...,rec_N}$ for a given user $id$ is represented as
$\{<id,rec_1>,<id,rec_2>,...,<id,rec_N>\}$ and $N$ is the number of
recommendations produced by the model. Based on the set of
recommendations and the hidden item for all the session in the test
set, we measure Recall, Precision and the F1 metric~\cite{yang:99,sarwar:00}:

\begin{center}
$Recall = \frac{|Hidden \cap Rec|}{|Hidden|}$, $Precision = \frac{|Hidden \cap Rec|}{|Rec|}$,
\end{center}

\begin{center}
$F1 = \frac{2 \times Recall \times Precision}{Recall + Precision}$.
\end{center}

Recall corresponds to the proportion of relevant recommendations.
Precision gives us the quality of each
individual recommendation. F1 is a measure that combines Recall and Precision with an equal
weight. It ranges from 0 to 1 and higher values indicate better
recommendations. Global recall, precision and F1 are obtained by
averaging individual test user values.

For the recommendation models based on association rules (AR), we
chose a minimum support value trying to keep at least 50\% of the items of
the data sets for building the models. The minimum confidence values were defined as being
the support value of the third most frequent item.

\subsection{Single Dimension}
\label{subsec:res}
\begin{figure*}[!ht]
\centering
\subfigure[CF in \textit{Listener} data set.]{\includegraphics[width=5.8cm]{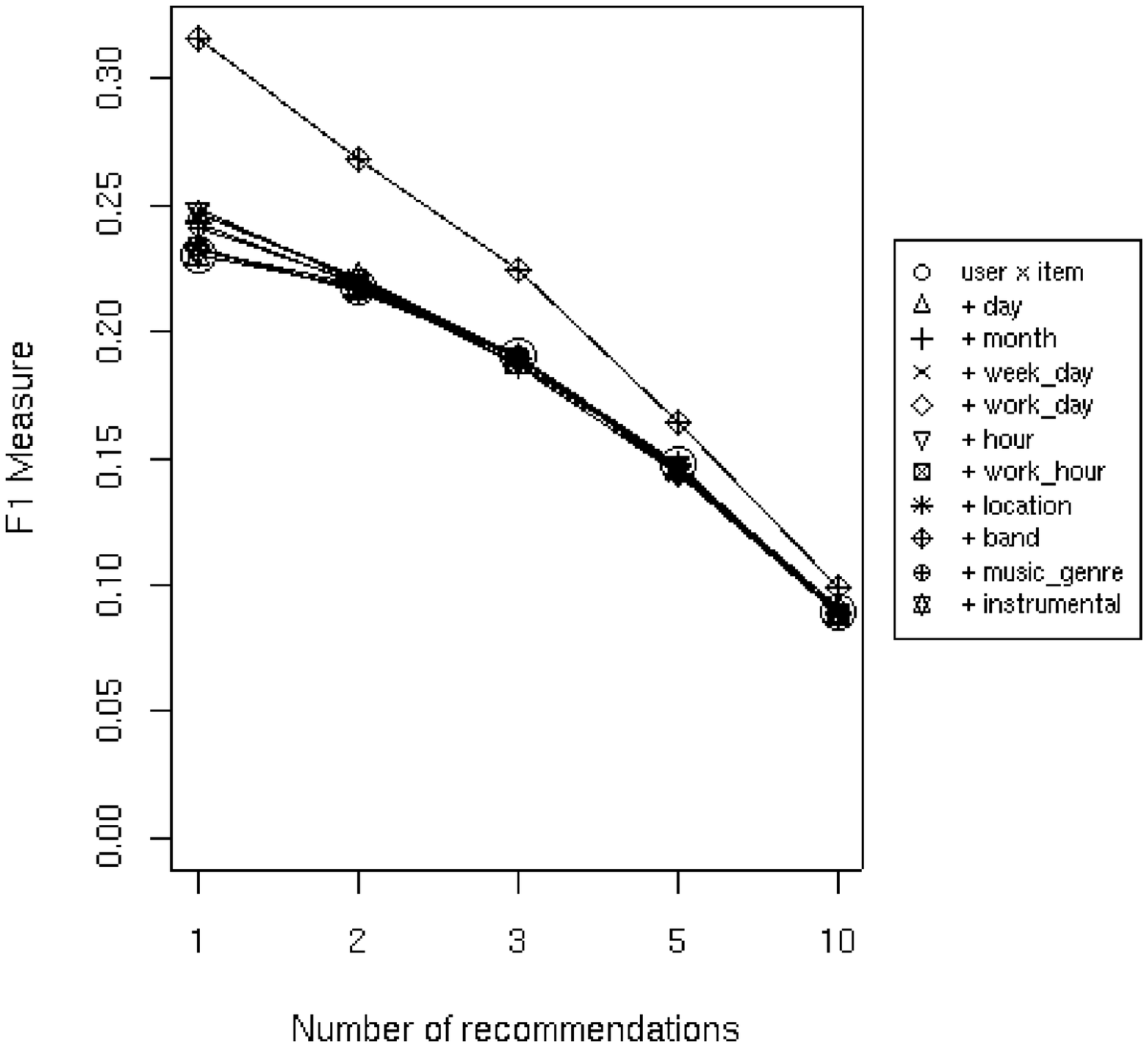}}
\subfigure[CF in \textit{Playlist} data set.]{\includegraphics[width=5.8cm]{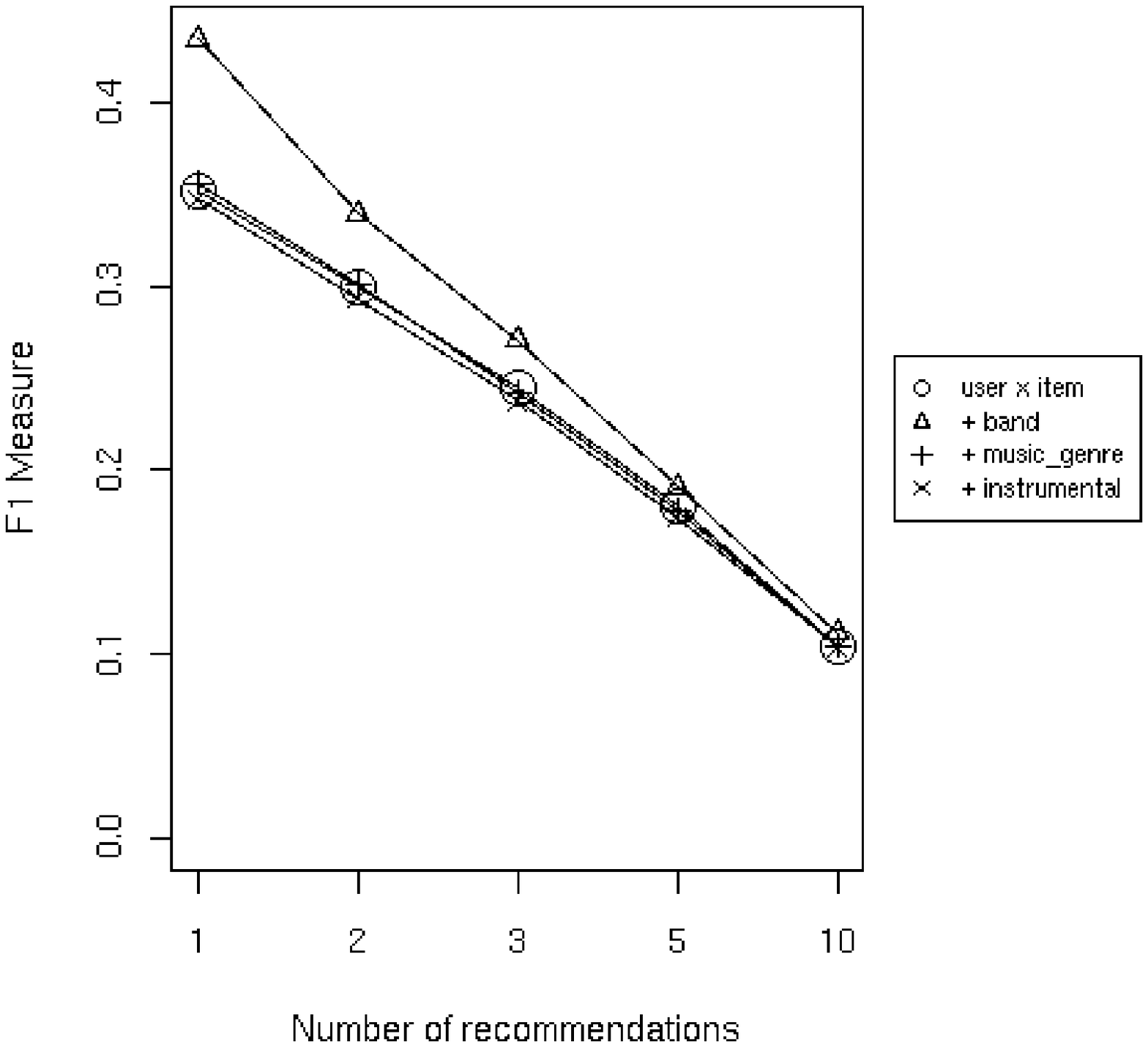}}
\subfigure[CF in \textit{Entree} data set.]{\includegraphics[width=5.8cm]{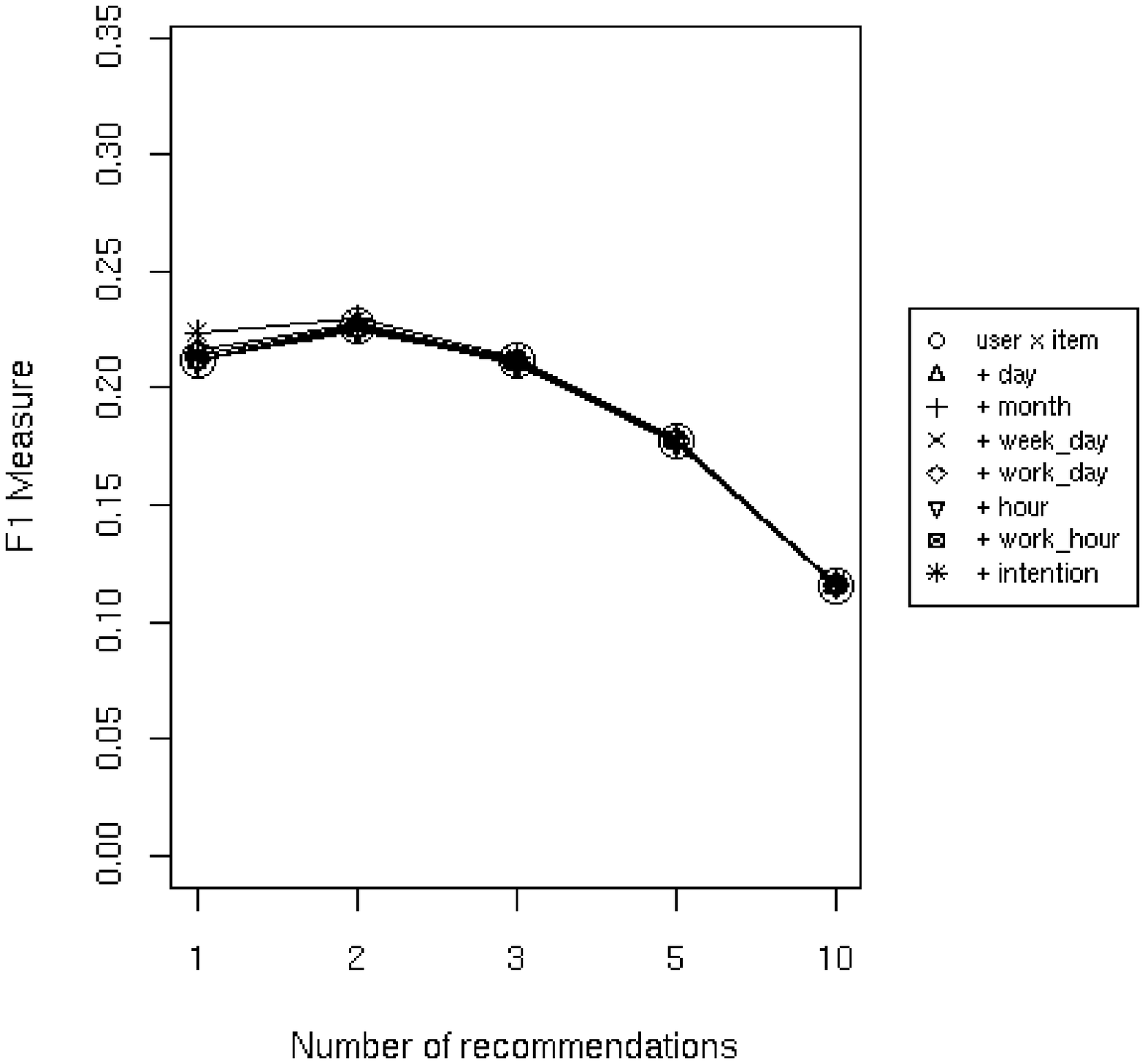}}
\subfigure[AR in \textit{Listener} data set.]{\includegraphics[width=5.8cm]{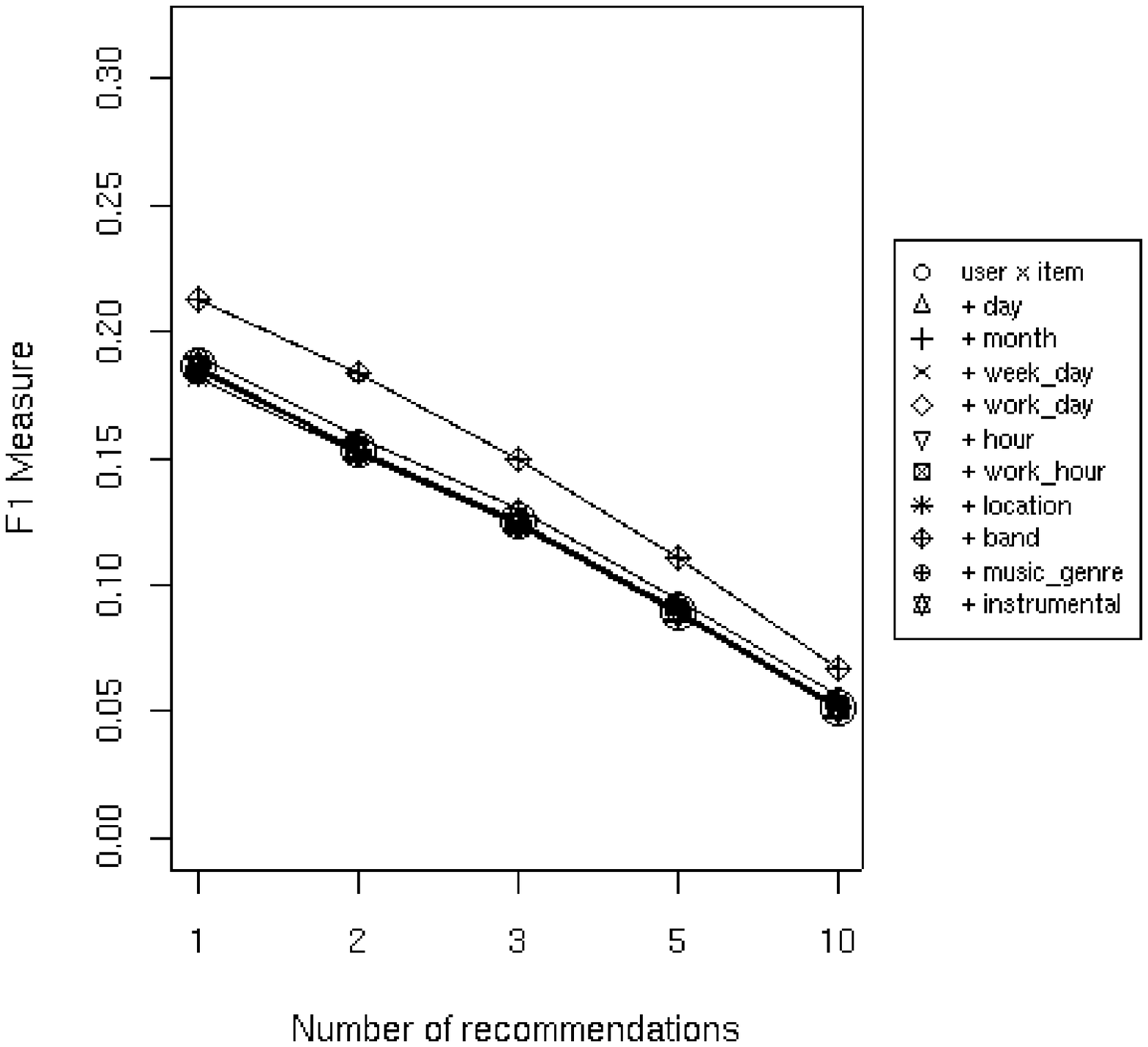}}
\subfigure[AR in \textit{Playlist} data set.]{\includegraphics[width=5.8cm]{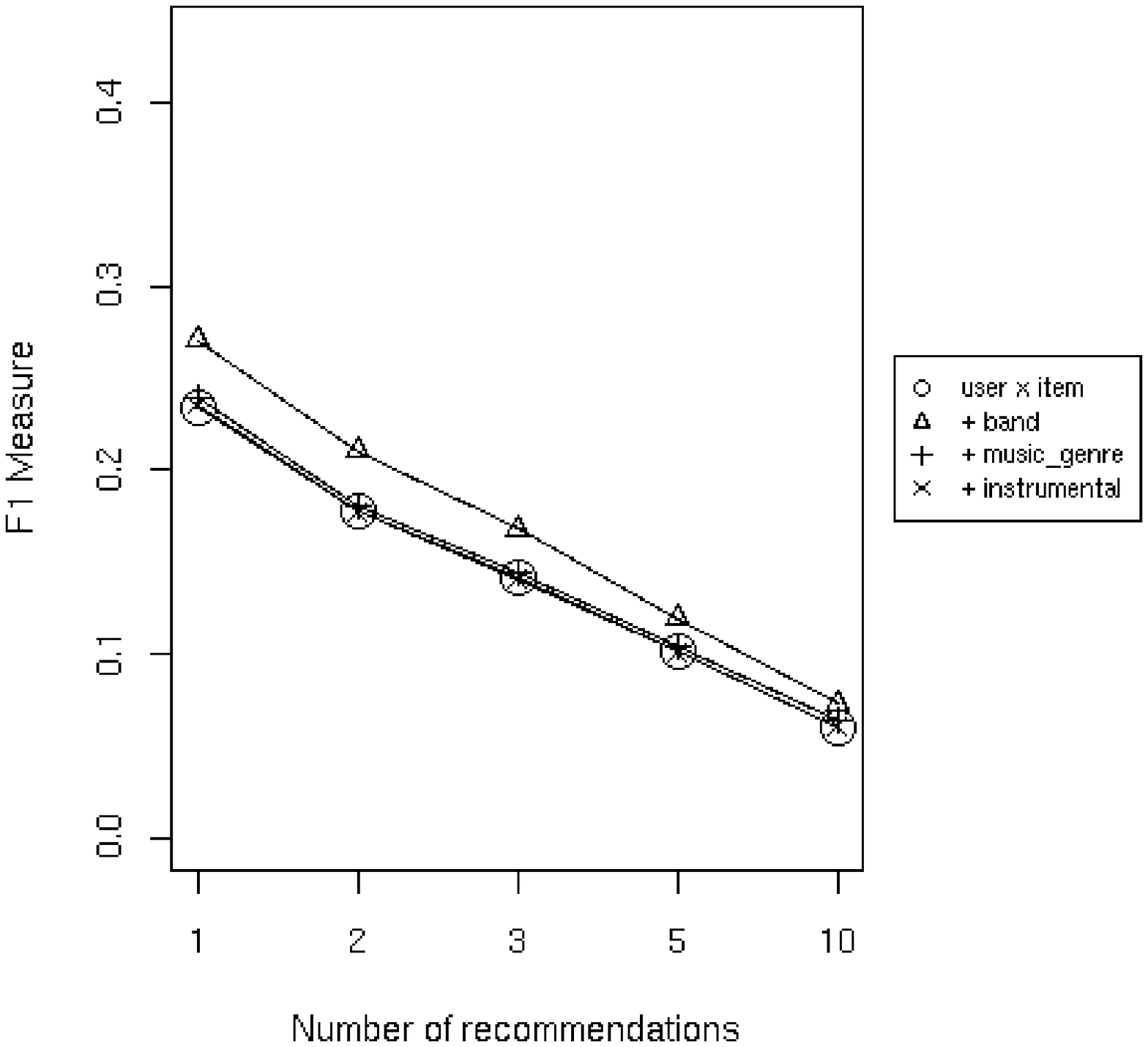}}
\subfigure[AR in \textit{Entree} data set.]{\includegraphics[width=5.8cm]{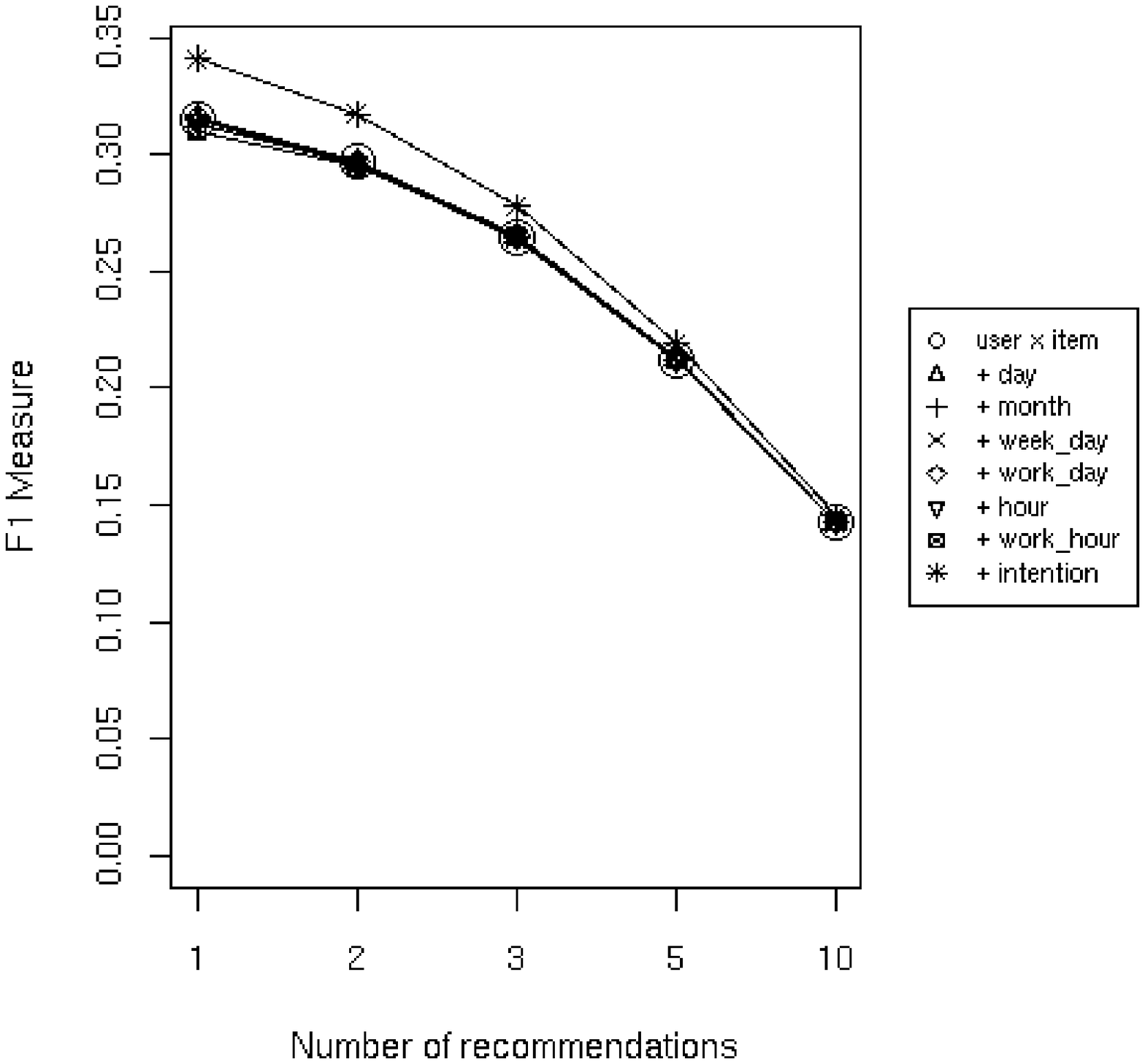}}
\caption{F1 metric for \textit{Listener}, \textit{Playlist} and \textit{Entree} data sets.}
\label{fig:f1_all}
\end{figure*}

Here we compare the results of the two algorithms using the traditional model
(user $\times$ item) and with \textbf{DaVI}, applied separately to each
contextual dimension presented in Table~\ref{table:dimensions}. The charts in
Figure~\ref{fig:f1_all} plot the F1 measure.

\subsubsection{Item-Based Collaborative Filtering}
Our results show that \textbf{DaVI} improves item-based CF predictive performance
when there is a rich contextual dimension. We can observe this with the
dimension \textit{band} in \textit{Listener} and \textit{Playlist} data sets
(Figure~\ref{fig:f1_all} (a) and (b)). In Figure~\ref{fig:f1_all} (a) the dimension \textit{band} yields a
maximal value of 0.31 (top 1). This value represents an F1 average gain
of 34\% compared to the value of F1 without applying \textbf{DaVI}.
In Figure~\ref{fig:f1_all} (b) \textit{band} provides a maximal value of 0.43 (top 1).
This value represents a gain of 24\%. In \textit{Entree} data set (Figure~\ref{fig:f1_all} (c)), the only expressive gain using contextual information was obtained with the dimension \textit{intention}, F1 value of 0.22 (top 1). This value represents a small gain of 5\%, which means that the CF model worked worse with \textit{intention} than \textit{band}, used in the former data sets. Additionally, we can see that the highest gains were obtained using dimensions collected from databases of the Web sites. The gains using dimensions pre-processed from Web access logs were very small. The values show average gains around 1.3\%, which means that the context inferred from Web access logs is not so rich in information.

\subsubsection{Association Rules}
Considering the association rules (AR) technique, our results also show that \textbf{DaVI} improves the accuracy of the recommendation models. In Figure~\ref{fig:f1_all} (d), we have a maximum value of 0.21 (top 1 with \textit{band}).
This value represents a gain of 14.5\%
compared to the value of F1 without any contextual information. In Figure~\ref{fig:f1_all}
(e), the F1 measure for the dimension \textit{band} and the top 2 has
the maximal F1 gain, 23.5\%. With respect to the \textit{Entree} data set
(Figure~\ref{fig:f1_all} (f)) we have a maximal value
of 0.34 (top 1 with \textit{intention}). This value represents a gain of 9.6\%.
An interesting fact here is that contrarily to the other data sets, the \textit{Entree}
presented highest F1 values with the association rules technique then with the
item-based collaborative filtering.

\subsection{Multiple Dimensions}
\label{subsec:res2}
So far, we applied \textbf{DaVI} to one contextual dimension at a time. However,
it may be applied to several dimensions.  We consider two different scenarios.
The first one (called \emph{all together}) simply applies the method to all
dimensions presented in Table~\ref{table:dimensions} simultaneously. The
 second scenario (called \emph{forward selection}) uses a sequential forward
 selection algorithm~\cite{jain:97}, on the training data set, to select the best combination of dimensions that will be used to make recommendations.
 The algorithm starts from an empty set and sequentially adds the dimension $d$ that results in the highest objective
function $F(D+d)$ on a validation data set, when combined with the dimensions $D$ that have already been
selected.

Given that there are no other methods that combine several contextual dimensions
for Top-$N$ recommendation, we compared our method to an adaptation of the
Combined Reduction approach~\cite{adomavicius:05a} for this task. Briefly, this
approach uses the values of the context/dimension as labels for segmenting Web
accesses and was originally developed for rating estimation. It consists of the
following two phases. First, using the training data, a recommendation method is
run for each contextual segment (e.g. accesses on Mondays would be a segment) to
determine which ones outperform the traditional model (using only user-item
information). Second, taking into account the context of the active session, we
choose the best contextual model to make the recommendation. Here the best model
is the one which has the highest F1 value. Here, we have adapted it for the
Top-$N$ recommender algorithms presented in Sections~\ref{subsec:cf} and~\ref{subsec:ar}.

As baselines, we have used the traditional user-item approach and also the
results of the best individual dimension (called
\emph{best context}), according to the previous experiments.

In Table~\ref{table:comp}, the results for $N=1$ show us that \textbf{DaVI}, using the best
dimension, has F1 values equal or higher than other \textbf{DaVI} scenarios. The only exception is the \textit{Listener} data set
with the CF model, where the best is \textbf{DaVI} using all dimensions. The Combined Reduction approach has values equal to and better than \textbf{DaVI} (best context), respectively, in \textit{Listener} and \textit{Playlist} data sets with the AR model. In Table~\ref{table:comp}, the symbol ``-'' means
that the algorithm timed-out.

\begin{table}[!ht]
\caption{F1 measure for Top-1 recommendations} \label{table:comp} \footnotesize
\centering                          
\begin{tabular}{p {4.1 cm } p {1 cm } p {1 cm } p {1 cm }}
\hline
 & \multicolumn{3}{c}{\textbf{CF}}\\
\multicolumn{1}{c}{\textbf{Methods}} & \textit{Listener} & \textit{Playlist} & \textit{Entree}\\
\hline \hline
user $\times$ item & \multicolumn{1}{c}{0.230} & \multicolumn{1}{c}{0.351} & \multicolumn{1}{c}{0.211} \\
\textbf{DaVI} (best context) & \multicolumn{1}{c}{0.315} & \multicolumn{1}{c}{\textbf{0.434}} & \multicolumn{1}{c}{\textbf{0.225}} \\
\textbf{DaVI} (forward selection) & \multicolumn{1}{c}{0.311} & \multicolumn{1}{c}{0.416} & \multicolumn{1}{c}{0.210} \\
\textbf{DaVI} (all together) & \multicolumn{1}{c}{\textbf{0.317}} & \multicolumn{1}{c}{0.429} & \multicolumn{1}{c}{\textbf{0.225}}\\
Combined Reduction & \multicolumn{1}{c}{0.225} & \multicolumn{1}{c}{0.351} & \multicolumn{1}{c}{0.212}\\
\hline
 & \multicolumn{3}{c}{\textbf{AR}} \\
\hline
user $\times$ item &  \multicolumn{1}{c}{0.186} & \multicolumn{1}{c}{0.234} & \multicolumn{1}{c}{0.315} \\
\textbf{DaVI} (best context) & \multicolumn{1}{c}{\textbf{0.213}} & \multicolumn{1}{c}{\underline{0.270}} & \multicolumn{1}{c}{\textbf{0.341}} \\
\textbf{DaVI} (forward selection) & \multicolumn{1}{c}{0.203} & \multicolumn{1}{c}{0.261} & \multicolumn{1}{c}{\textbf{0.341}} \\
\textbf{DaVI} (all together) & \multicolumn{1}{c}{-} & \multicolumn{1}{c}{0.268} & \multicolumn{1}{c}{0.336} \\
Combined Reduction & \multicolumn{1}{c}{\textbf{0.213}} & \multicolumn{1}{c}{\textbf{0.280}} & \multicolumn{1}{c}{0.309} \\
\hline
\end{tabular}
\end{table}

\section{Conclusions and Future Work}
\label{sec:conc}
In this paper we presented a direct approach, called \textbf{DaVI}, that enables existing
recommender systems to take advantage of contextual information as virtual items. We discussed the results obtained using two recommendation
techniques, item-based collaborative filtering and association ru\-les. Using \textbf{DaVI} with rich contextual information has revealed a great potential to improve the accuracy of recommender systems. However identifying rich contextual dimensions is not an easy task.

We have also compared different settings using the \textbf{DaVI} approach
(best dimension, forward selection and all dimensions) with a more sophisticated
Combined Reduction approach. Next, we will improve this empirical study and propose a method to identify rich contextual information from Web sites that can be used with \textbf{DaVI}.

\section{Acknowledgments}
Fund. Ci\^{e}ncia e Tecnologia (PhD Grant SFRH/BD/22516\\/2005) and grant QREN-AdI Palco3.0/3121 PONORTE.

\bibliographystyle{abbrv}
\bibliography{references}

\begin{thebibliography}{10}

\bibitem{adomavicius:05a}
G.~Adomavicius, R.~Sankaranarayanan, S.~Sen, and A.~Tuzhilin.
\newblock Incorporating contextual information in recommender systems using a
  multidimensional approach.
\newblock {\em ACM Transactions on Information Systems}, 23(1):103--145, 2005.

\bibitem{agrawal:94}
R.~Agrawal and R.~Srikant.
\newblock Fast algorithms for mining association rules.
\newblock In {\em Proceedings of Twentieth International Conference on Very
  Large Data Bases}, pages 487--499, 1994.

\bibitem{anand:03}
S.~S. Anand and B.~Mobasher.
\newblock Intelligent techniques for web personalization.
\newblock In {\em Intelligent Techniques for Web Personalization (ITWP 2003),
  LNCS 3169}, pages 1--36, 2003.

\bibitem{breese:98}
J.~S. Breese, D.~Heckerman, and C.~M. Kadie.
\newblock Empirical analysis of predictive algorithms for collaborative
  filtering.
\newblock In {\em Proceedings of the 14th Conference on Uncertainty in
  Artificial Intelligence}, pages 43--52, 1998.

\bibitem{dey:01}
A.~K. Dey.
\newblock Understanding and using context.
\newblock {\em Personal Ubiquitous Computing}, 5(1):4--7, 2001.

\bibitem{dey:01a}
A.~K. Dey, G.~D. Abowd, and D.~Salber.
\newblock A conceptual framework and a toolkit for supporting the rapid
  prototyping of context-aware applications.
\newblock {\em Human-Computer Interaction}, 16(2):97--166, 2001.

\bibitem{domingues:07}
M.~A. Domingues, A.~M. Jorge, C.~Soares, J.~P. Leal, and P.~Machado.
\newblock A data warehouse for web intelligence.
\newblock In {\em Proceedings of the 13th Portuguese Conf. on Artificial
  Intelligence}, pages 487--499, 2007.

\bibitem{gorgoglione:06}
M.~Gorgoglione, C.~Palmisano, and A.~Tuzhilin.
\newblock Personalization in context: Does context matter when building
  personalized customer models?
\newblock In {\em ICDM '06: Proceedings of the Sixth International Conference
  on Data Mining}, pages 222--231, 2006.

\bibitem{jain:97}
A.~Jain and D.~Zongker.
\newblock Feature selection: evaluation, application, and small sample
  performance.
\newblock {\em IEEE Transactions on Pattern Analysis and Machine Intelligence},
  19(2):153--158, Feb 1997.

\bibitem{karypis:01}
G.~Karypis.
\newblock Evaluation of item-based top-n recommendation algorithms.
\newblock In {\em CIKM'01: Proceedings of the 10th International Conference on
  Information and Knowledge Management}, pages 247--254, 2001.

\bibitem{palmisano:08}
C.~Palmisano, A.~Tuzhilin, and M.~Gorgoglione.
\newblock Using context to improve predictive modeling of customers in
  personalization applications.
\newblock {\em IEEE Trans. on Knowl. and Data Engineering}, 20(11):1535--1549,
  2008.

\bibitem{sarwar:00}
B.~Sarwar, G.~Karypis, J.~Konstan, and J.~Riedl.
\newblock Analysis of recommendation algorithms for e-commerce.
\newblock In {\em Proceedings of the 2nd ACM Conf. on Electronic Commerce},
  pages 158--167, 2000.

\bibitem{yang:99}
Y.~Yang and X.~Liu.
\newblock A re-examination of text categorization methods.
\newblock In {\em Proceedings of the 22nd annual international ACM SIGIR
  Conference on Research and Development in Information Retrieval (SIGIR'99)},
  pages 42--49, 1999.

\end{thebibliography}

\end{document}